\documentclass[10pt,twocolumn,letterpaper]{article}

\usepackage{wacv}
\usepackage{times}
\usepackage{epsfig}
\usepackage{graphicx}
\usepackage{amsmath}
\usepackage{amssymb}
\usepackage{multicol}
\usepackage{caption}
\usepackage{subcaption}
\usepackage[keeplastbox]{flushend}

\wacvfinalcopy 

\def\wacvPaperID{****} 

\ifwacvfinal\fi
\begin{document}

\title{Leveraging Localization for Multi-camera Association}

\author{Zhongang Cai \hspace{2cm} Cunjun Yu \hspace{2cm} Junzhe Zhang  \\ Jiawei Ren \hspace{2cm} Haiyu Zhao \\
Innova \\
{\tt\small cai.zhongang@gmail.com}
}

\maketitle
\ifwacvfinal\thispagestyle{empty}\fi

\begin{abstract}
We present McAssoc, a deep learning approach to the association of detection bounding boxes in different views of a multi-camera system. The vast majority of the academia has been developing single-camera computer vision algorithms, however, little research attention has been directed to incorporating them into a multi-camera system. In this paper, we designed a 3-branch architecture that leverages direct association and additional cross localization information. A new metric, image-pair association accuracy (IPAA) is designed specifically for performance evaluation of cross-camera detection association. We show in the experiments that localization information is critical to successful cross-camera association, especially when similar-looking objects are present. This paper is an experimental work prior to MessyTable \cite{cai2020messytable}, which is a large-scale benchmark for instance association in mutliple cameras.
\end{abstract}

\section{Introduction}

The rapid advancements in convolutional neural networks (CNNs) have transformed the landscape of computer vision research, leading to a boom of various high performing applications that takes advantage of deep learning's exceptional capability in complicated interpretation and robustness against noise. 

Despite the unprecedented performance, most works have assumed a single camera set-up and little research attention has been directed to building a multi-camera system upon the existing works. 
Multi-camera detection systems consist of synchronized cameras with overlapping views of the same scene. Multi-camera systems are prevalent in real life and extending a single-camera system to multi-camera system has significant advantages. For example, an indoor surveillance system can benefit from having multiple view angles to handle severe occlusion when the area gets crowded; an autonomous car can expand its field of view by having several cameras mounted on the top; an unmanned factory can strategically place a top camera to determine an object's location on the conveyor belt, while fusing with a side camera that detects a label that is not visible from the top. In a nutshell, a multiple-camera system alleviates the effects of clustering, expands the field of view and allows cross-camera fusion of high-level information.

Hence, we propose McAssoc, to build upon previous research on the multi-camera systems but with a focus on detection association. In this paper, we assumed object detection models have been run for each of the synchronized images, and McAssoc is placed at the end of these detection pipelines to associate bounding boxes across different views. 

In the past, bounding boxes are associated based on the appearance features or location of the target objects. However, appearance features alone does not handle identical objects; location information is prone to corner cases and noisy detection. Hence, we propose a McAssoc module that takes in inputs that combine both feature and location information. The McAssoc module is designed to be versatile, able to give different outputs with different heads. On top of the McAssoc module, we design a 3-branch architecture that utilizes the versatility of McAssoc modules for a highly accurate and robust association of detection boxes.

Since the topic is less well-represented in the academia, we implement a reasonably effective classical approach that is commonly used to handle multi-camera object association tasks based on homographic projection. We have also implemented a version of McAssoc module that takes in only the feature information to show location information is important.

In addition, we collect a bounding box detection association dataset to validate our arguments. The dataset is designed to include challenging scenes where many similar- or identical-looking objects are present.

\section{Related Works}
Many computer vision tasks are essentially association. For example, optical flow \cite{ref:flownet} is the association of pixels across frames to predict the movement of the camera or the scene; similarly, tracking also uses temporal information and can be seen as an object-level association from one frame to another; stereo vision, however, does not perform association over the temporal but the spatial axes, and associate pixels across cameras. There exists a gap where little research attention has been shed on the object-level spatial association, that is, multi-camera detection association.

In this section, we evaluate relevant research works, especially those which have a multi-camera component.

\subsection{Monocular Detection}
With recent advances in the deep neural network, the accuracy of monocular detection has been drastically improved. Both single-shot detector \cite{ref:ssd,ref:yolo,ref:retinanet} and two-stage detector\cite{ref:frcnn,ref:cascadercnn} offer decent results in real applications. The model trained with properly labeled data is capable of locating target objects with a class label at one shot. However, all these methods are designed for a single-camera setup. One major drawback is the model has no clue where to find the target objects when heavy occlusions are imposed. This is caused by the inherent lack of global information with such a single-camera setup, which highlights the need for a multi-camera detection system.

\subsection{Feature-Based Multi-Camera Detection}
Great efforts have been put into extending monocular detection to multi-camera detection. One main approach to fusing information captured from a different perspective is to compare the appearance of target objects bound by detection boxes. Features of target objects can be either extracted using a classical algorithm\cite{ref:tunnel}, or a deep neural network\cite{ref:alexnet}. By comparing a pair of features from different viewpoints using a binary classifier, one can determine whether they belong to the same target object.

An efficient way of feature matching or fusion is crucial in tasks with multiple inputs since it determines the pairing results for a feature-based approach. One vanilla approach would be transforming the feature map into feature vectors followed by a concatenation operation for two vectors \cite{ref:goturn}. However, to find out the stronger interrelations between features, a novel method with a new correlation layer proposed in \cite{ref:geometric} is then developed to produce a finer matching.

\subsection{Location-Based Probability Occupancy Map}
Distance between feature points of target objects can be strong evidence indicating the pairing relationship between captured images of target objects. In \cite{ref:dmcpd}, a probability occupancy map (POM) is established to estimate the presences of people in the scene to assist the tracking and detection of pedestrians with a multi-camera setup. However, this series of work does not give explicit association results between bounding boxes across camera views.

\subsection{Tracking}
A similar task of multi-camera detection would be tracking. Both of these two tasks require multiple inputs for output purpose. With the wide usage of deep neural network in computer vision, tracking algorithm evolves from relative simple architecture \cite{ref:goturn} to a much more advanced implementation \cite{ref:siameserpn} recently. However, there are two major differences between tracking tasks and multi-camera detection, which limits the direct transfer of method from tracking. While the temporal information is the dominant factor people mainly cope with in tracking tasks, spatial information instead in multi-camera detection triggers more attention from researchers because images are taken from different points of view synchronously. The other major difference is in most tracking scenarios, the scene or the background is not changing except the target objects are moving.

\subsection{Relative Extrinsic Parameters between Multiple Views}
Relative camera pose estimation is a task that takes in a pair of images and outputs the relative extrinsic parameters between the two respective cameras by which the images are captured. Previous studies \cite{ref:relative_camera, ref:posenet, ref:vsnet} have all used a similar architecture that involves a Siamese feature extraction stage, followed by feature matching and global regression. Multi-sensor calibration \cite{ref:regnet} and visual servoing \cite{ref:vs} are also applications of such architecture for they require estimations of relative extrinsic parameters. Although McAssoc does not require accurate relative extrinsic parameters, understanding the relative camera poses is crucial for finding the correspondences of objects in two views.

\subsection{Stereo Vision}
Stereo vision uses two cameras and estimates the depth of each pixel by finding correspondences of pixels in two views. However, the two cameras are usually placed side by side to capture subtly different images \cite{ref:kitti}. However, this property is not applicable in the multi-camera association task because the cameras are usually placed at vastly different angles and positions. 

\section{Neural Network}
\label{sec:nn}

\subsection{McAssoc Module}
\begin{figure*}[ht]
\centering
  \includegraphics[width=0.8\textwidth]{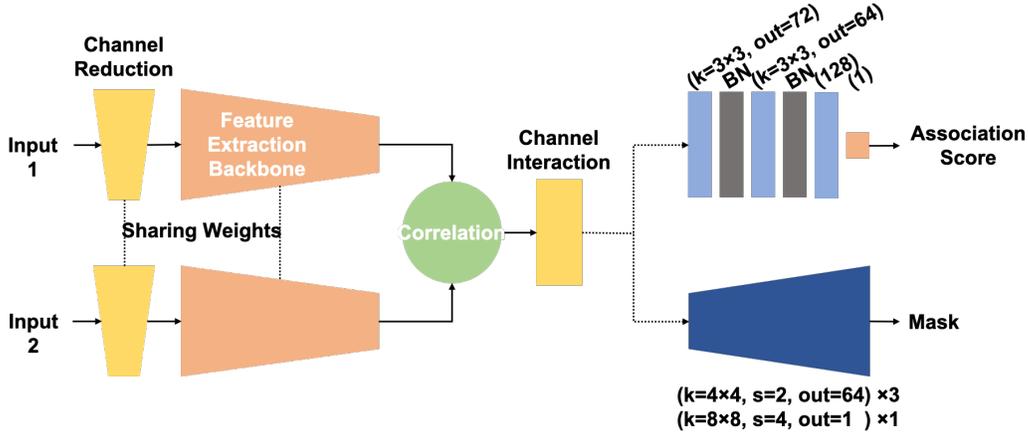}
\caption{McAssoc Module with Two Possible Heads}
\label{fig:module}
\end{figure*}

We firstly design a versatile McAssoc module that has two possible heads to choose from for mask or association score computation. Note any module instance has only one of the two heads, not both. The association score represents how likely two instances belong to the same object.
Although a McAssoc module with association score head is by itself a decent design for association prediction, we topped it with several other McAssoc modules with different heads to build the 3-branch architecture that gives a highly accurate and robust prediction.

We designed the input to have an RGB image and an attention mask of the same size (both resized to $224 \times 224$) to encode both feature and location information. 

The RGB image serves two purposes: firstly, similar to a camera pose estimation task \cite{ref:relative_camera, ref:posenet}, pixel correspondences can be found from two images to find the spatial relationship between the two views; secondly, the full image provides feature information in the neighborhood of the target object, which extends feature-based association to include the immediate surrounding in addition to the target object itself.

The attention mask is inspired by Box Attention\cite{ref:box}, the values of pixels that are outside the bounding box are set to 0 whereas those inside are 1. The attention mask \textit{implicitly} provides location information. The full RGB images enable interpretation of the relative camera poses that allow bounding boxes in different views, which are marked by the attention box, to be related spatially. Note that no camera extrinsic parameters are passed to the network; the model interprets the relative camera poses internally.

A Siamese architecture \cite{ref:siamese} is used in the feature extraction stage for the two inputs. Each input firstly goes through the channel reduction layer, a convolutional layer that reduces the channel number from 4 to 3. Hence, the input is then compatible with any of the mainstream backbones (we use the light-weight ResNet-18 in this paper; details can be found in \ref{sec:er}). The weights of the channel reduction layer and the backbones for two inputs are shared.

The matching of two feature maps, $f_A, f_B \in \mathbb{R}^{h\times{w}\times{d}}$, from the feature extractor is done by a correlation layer \cite{ref:geometric} which is formulated as: 
$$ c_{AB}(i,j,k)=\mathbf{f_{B(i,j)}}^T \mathbf{f_{A(i_k,j_k)}} $$
To put in words, the column of size $d$ at $(i,j)$ of feature map $f_A$ is taken out to be a vector $\mathbf{f_A}$. Similarly, a vector $\mathbf{f_B}$ at $(i_k,j_k)$ of the feature map $f_B$ is taken out. The dot product of these two vectors are placed at $(i,j,k)$ of the output feature map $c_{AB}$ of size $(h, w, h \times w)$. In detail, for each $\mathbf{f_A}$, it loops through the entire $f_B$ (at each $(i_k, j_k)$), obtaining total $h \times w$ dot product values and place these values at the same row and column indices $(i, j)$ of the $c_{AB}$. We implement a vectorized, much faster implementation that in our source code.

A channel interaction layer follows the correlation layer to allow further interaction within the correlated feature map as well as control the channel depth to 64.

Two types of heads can be attached afterward for different purposes. 
\begin{itemize}
    \item Association head. The association head contains two convolution layers with batch normalization, flattening and two fully connected layers. ReLU is used as the activation function except for the last layer where sigmoid is used. It outputs a single association score.
    \item Mask head. The mask head contains 4 transpose convolution layers (referred by many as deconvolution layers) that upsample the feature map to match the size of the original input image size with depth 1. Sigmoid activation is applied before output, which is the predicted attention mask. Note for mask head, the attention mask in input 2 is set as all-zero.
\end{itemize}

For both heads, binary cross entropy loss is used (for mask head, the loss is used at pixel level):
$$Loss_{BCE} = -\frac{1}{N}\sum_{i=1}^{N}y_i\cdot\log(\hat{y_i}) + (1-y_i)\cdot\log(1-\hat{y_i})$$

\subsection{3-branch Architecture}

\begin{figure*}[ht]
\centering
  \includegraphics[width=0.8\textwidth]{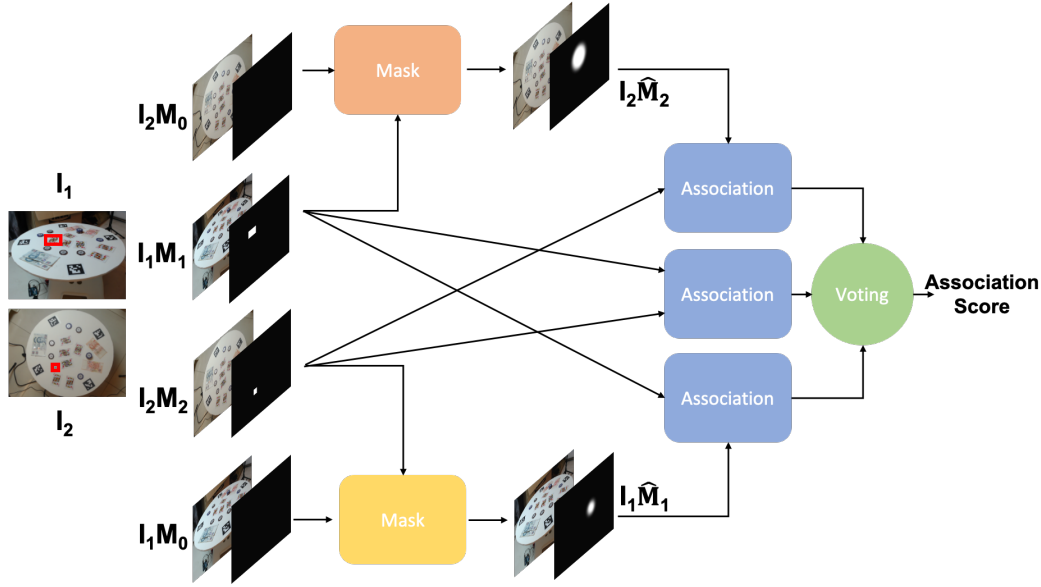}
\caption{3-Branch Architecture. Images with attention mask are fed into three different association modules to achieve better association prediction. The association score is obtained by utilizing all three branches.
}
\label{fig:3branch}
\end{figure*}

We also take one step further to design a 3-branch architecture, that comprises of two McAssoc modules with mask head (referred as mask module below) and three McAssoc modules with association score head (referred below as association module). 

The central association module is easy to understand as it takes in two inputs, each has a full image and an attention mask, to predict if the two objects are the same instance. 

The other two association modules take advantage of the mask module's ability to provide cross-location information (takes in an object's location in one image, predicts the object's location in another image). We argue that localization information is critical especially when similar-looking objects are present.

In Figure \ref{fig:3branch}, take the yellow mask module for example, it takes in two inputs: first, image 2 ($I_{2}$) and the attention mask of image 2 ($M_{2}$); second, image 1 and a all-zero attention mask $M_{0}$. The yellow mask module thus predicts the attention mask of image 1 ($\hat{M}_{1}$). Then $I_{1}$ is combined with the predicted mask $\hat{M}_{1}$ as an input to one of the association modules (the one at the bottom). This association module also takes in Image 1 and the true attention mask ($I_{1}M_{1}$). The idea is straightforward: if the items in two images are the same instance, the predicted mask should be close to the true mask. In addition, the association module is trained with the case where two inputs are generated from the same camera view so it can handle such a situation.

The voting layer can simply take the average or the maximum of the three association scores coming out of the association modules. However, we use 3 layers of fully connected layers to learn a strategy to assign weights to the scores.

\section{Dataset}
\label{sec:dataset}

Multi-camera detection association is not sufficiently studied. This can be partially attributed to a lack of large-scale datasets \cite{ref:dmcpd}. The existing datasets are small and not designed for object detection association \cite{ref:pets}. For example, several datasets for probabilistic occupancy map problems \cite{ref:wildtrack} provides ground truth occupancy maps \footnote{Occupancy maps are virtual grids on the ground and each grid contains a binary value to indicate if there is a person present in the grid.} but no association labels (if one bounding boxes is associated with another one in another camera view) can be extracted. Hence, we designed a rig to collect a dataset for training. 

\subsection{Design}
The setup consists of a turntable that is driven by an electrical motor and rotate at a constant rate. Three cameras are installed around the turntable, one view from the top and the rest at the side (left and right). A \textit{scene} refers to an arrangement of items on the turntable surface. 20 sets of synchronized images (each set consists of 3 synchronized images) are taken at a constant interval during one revolution of the turntable for each scene. We collected 50 scenes in total.

It is important to note that although only 3 cameras are used, much more camera poses can be utilized: as the turntable rotates, the camera poses relative to the turntable changes, hence, a total 60 different camera poses are available for cross-camera association.

Three types of items are used: casino chips, playing cards, and cash. These types of items are selected because a large number of similar features within each type can be obtained through different combinations of individual items. For example, a stack of casino chips can contain an arbitrary number of chips of any values. An \textit{object} in the data set can be a stack of casino chips, a playing card or several pieces of overlapping cash notes.

The arrangement of items is designed to be challenging: identical (at least in one view) items are purposely placed in the same scene. For example, two stacks of the same combination of casino chips, two pieces of cash bill of the same value and so on. In addition, items are placed in dense clusters to allow significantly overlapping bounding boxes to add the confusion.

\subsection{Annotation}
The list of items used is known and not changed within one scene. We annotate all bounding boxes and assign each bounding box an index. 
The data is labeled in the following stages: 
1. We annotate all foreground objects with bounding boxes in all images. 2. We select an image from the scene in which all objects are visible. We refer to this image as the reference image. We assign index 1 to n to the bounding boxes in the reference image.
3. Bounding boxes in other images are assigned the same index as their corresponding box in the reference image.


\subsection{Example Generation}
\label{sec:data_amount}
For each bounding box in each image, a bounding box of the same index can be found in another image in the scene to form a positive example. However, the number of negative examples is very large as any bounding box with a different index can form a negative example with the current bounding box. Hence, we define \textit{data amount} as the number of negative examples found for each positive examples. If the data amount is more than $1\times$, the positive example is resampled to balance the number of negative examples.

\subsection{Noises}
\subsubsection{Imperfect bounding boxes}
Our annotators are trained to annotate tight-fitting bounding boxes. However, the actual detection bounding boxes are usually not fully overlapping with ground truths. The research community uses mIoU from 0.5 to 0.95 as the threshold to gauge the performance of detection models \cite{ref:coco}. As McAssoc is assumed to be used downstream of detection models, we also test manually adding in noises to the location and the dimension of the bounding boxes. We generate random imperfect bounding boxes from the original ground-truth given the IoU of the two boxes. Implementation can be found in our source code.

\subsubsection{False or missed detections}
Detection models can give false positive (the model predicts a bounding box but there is no ground truth object, a false detection) or false negative (the model does not predict bounding box but there is a ground truth object, a missed detection) results. 

To simulate false detection, we test adding in a random number of bounding boxes. The bounding box location and dimension are randomly sampled from the range for which the minimum and maximum is defined by the extremes of the distribution of the ground truth bounding boxes. Hence, the  'fake' bounding boxes are generated to be realistic as they do not appear at strange corners or have unusual dimensions.

To simulate missed detection, simply a random number of ground truth bounding boxes are removed from each image.

\section{Classical Approach}
As there is often a dominant 2D plane that can be found in our applications (such as a multi-camera surveillance system that monitors pedestrians on the flat road surface, or in our dataset, objects on a flat table surface), homographic projection is appropriate to transform bounding boxes from one camera to another. For systems that contain more than two cameras, one camera can be seen as the main camera and bounding boxes in the rest of the cameras' views can be projected to the main camera. In our setup, the top (bird view) camera is chosen to be the main camera for clearest visualization of the association results.

Projection of the entire bounding box in the sindexe view is both inappropriate and unnecessary because only the bottom edge can be regarded as "touching" the surface of interest (the tabletop), the rest of the bounding box, however, forms an elongated trapezium that does not have relevant physical meaning. We thus only consider the middle point of the bottom edge of the bounding box in the side view to be the key point to be projected. For the bounding box in the top view, the center is used as the key point.

The homographic projection is:
$$
s
\begin{bmatrix}
\mathbf{p'} \\
1
\end{bmatrix}
= H \times
\begin{bmatrix}
\mathbf{p} \\
1
\end{bmatrix}
$$
where $\mathbf{p}$ and $\mathbf{p'}$ are points (vectors consisting of x, y) before and after projection.

Hence, two lists of key points ($m$ top view key points and $n$ projected side view key points) can be matched as a maximum matching with minimum total cost problem, for which the cost is defined to be the pixel distance between two key points from two lists. 

The cost matrix is thus fomulated as:
$$C_{ij} = ||\mathbf{p_{i}} - \mathbf{p_{j}}||, C \in \mathbb{R}^{n \times m}$$

Kuhn-Munkres algorithm (also known as the Hungarian algorithm) is used. Note that for each object, a threshold $t$ is applied and all cost values higher than the threshold is invalidated before matching by setting to the same value higher than $t$ (for example, $t + 1$). After matching, a match with such an invalid cost is also rejected.

\subsection{Analysis}

The classical approach requires well-calibrated distortion parameters for the flat surface assumption to hold. In addition, the homography matrix should be computed online through key point matching algorithms such as ORB \cite{ref:orb} or SIFT \cite{ref:sift} or calibrated either beforehand. 
Computation of the homography matrix can introduce new errors due to noises and the presence of feature points that do not belong to the flat surface. In our implementation, the matrix is calibrated by the manual marking of corresponding points but this is a labor-intensive task and thus results in poor scalability. Also, as the cameras can be subjected to collision or shaking which requires recalibration.

Moreover, as the middle point of the bottom edge of the bounding box is used as the key point, it suffers from imperfect bounding box parameters (x and y for the location of the bounding box in the image and w and h for the dimensions). It is very common to have bounding boxes that do not tightly enclose the objects, leading to errors of the pixel coordinates of the key points.

Furthermore, as the side view has no depth information, the projected key point will never coincide with the top view key point (which is the center of the bounding box) but differ by at least the distance from the center to its closest edge even if the bounding box is tight. This property poses no impact on the global optimum matching as long as perfect detection can be assumed. However, it is very common to have false detection (associated with precision) and missed detection (association with recall), such corner cases can disturb the matching severely. 

In addition, the threshold $t$ has to be tuned. Too large the value results in poor matching accuracy, too small the value rejects correct matching. It is especially difficult to tune the value when the objects of interest have a large variety of sizes.

Very importantly, the classical method makes sense only if the main camera has its image plane parallel to the dominant 2D plane in the scene. Otherwise, the pixel distance cannot be compared to determine the closeness of the objects as the same pixel distance indicate larger physical distance further away from the camera.

The analysis is supported by experiments in Section \ref{sec:vs_classical}

\section{Experiments on Localization}
\label{sec:exp_local}
In Figure \ref{fig:3branch} and Section \ref{sec:nn}, the McAssoc modules with mask head (refer below as mask module) provide cross localization information. We conduct experiments on the mask module to evaluate its performance on location mask prediction.

Identical to its actual role in the 3-branch architecture, the mask module is given two sets of input, each consists of a full image (3-channel) and an attention image (1-channel and binary). The attention mask in output 2, however, has all values set to 0. 

As mentioned above, the attention mask indicates the object of interest. Hence, the mask module is tasked to predict the actual attention mask for the full image in output 2 that overlaps with the same object of interest in input 1.

\subsection{Training}

Similar to semantic segmentation task, the module gives a pixel-level prediction. However, since there is only one class, we train the module with a binary cross-entropy loss, supervised with the actual attention mask of the full image in input 2.

In addition, the attention area can be small (for example, a chip occupies typically 0.5\% of the total image area), we place a significantly larger weight on pixels with a positive label. Positive weight, $\mathbf{2\times10^4}$, is empirically optimal in this case.

\subsection{Results and Visualization}

We evaluate the quality of the predicted attention mask by computing its  intersection of union (IoU) with the ground truth mask. The ground truth mask is essentially a binary map where the bounding box is filled with one and the rest is filled with zero. An average IoU across all test examples is around 0.25.

\begin{figure}
  \begin{subfigure}[b]{\columnwidth}
    \centering\includegraphics[width=\columnwidth]{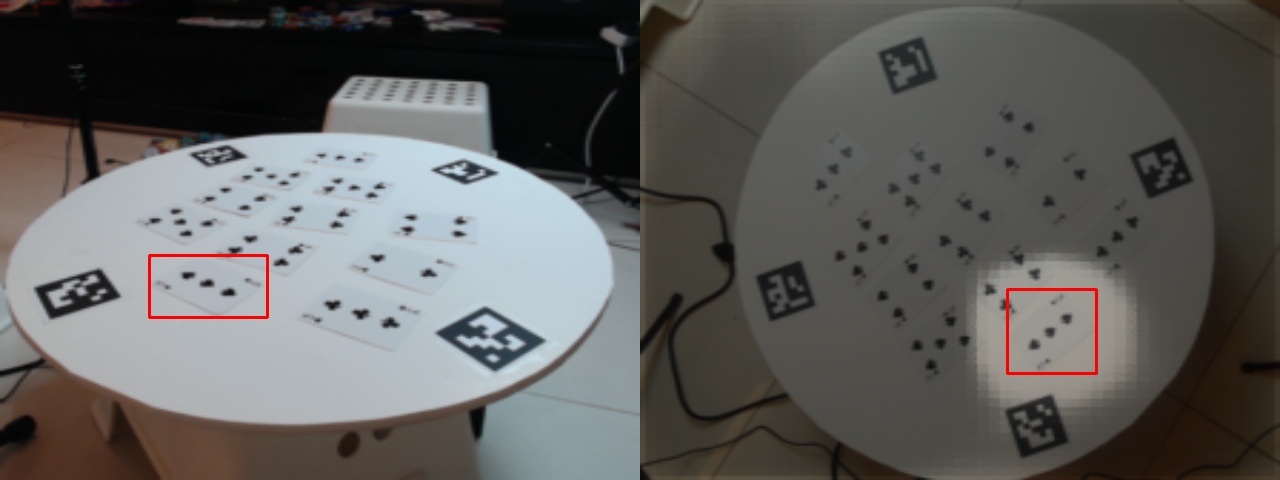}
    \caption{Attention mask predicted for a playing card.}
  \end{subfigure}
  \begin{subfigure}[b]{\columnwidth}
    \centering\includegraphics[width=\columnwidth]{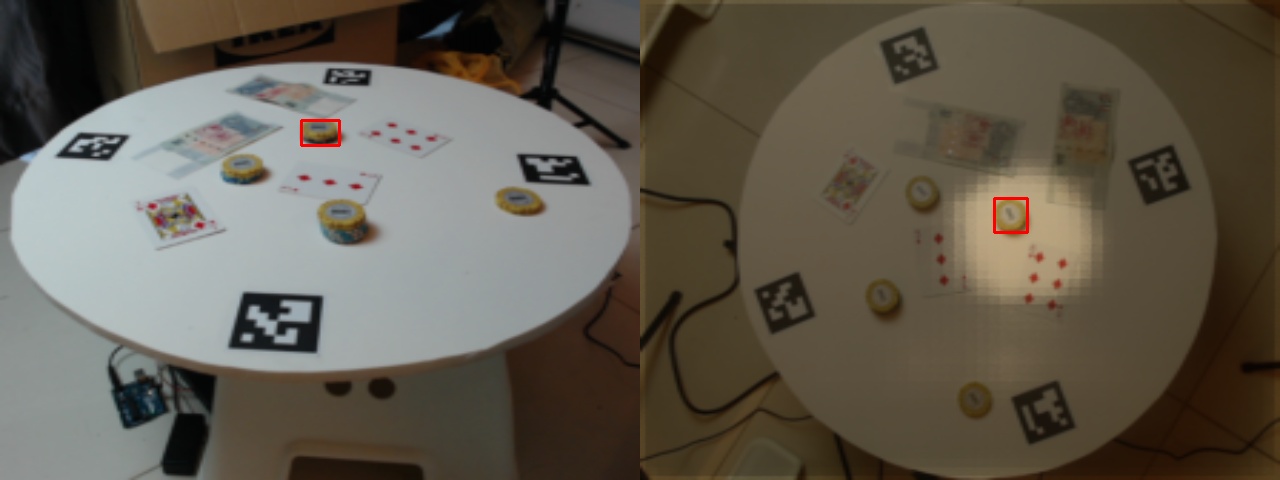}
    \subcaption{Attention mask predicted for a chip stack.}
  \end{subfigure}
  \begin{subfigure}[b]{\columnwidth}
    \centering
    \includegraphics[width=\columnwidth]{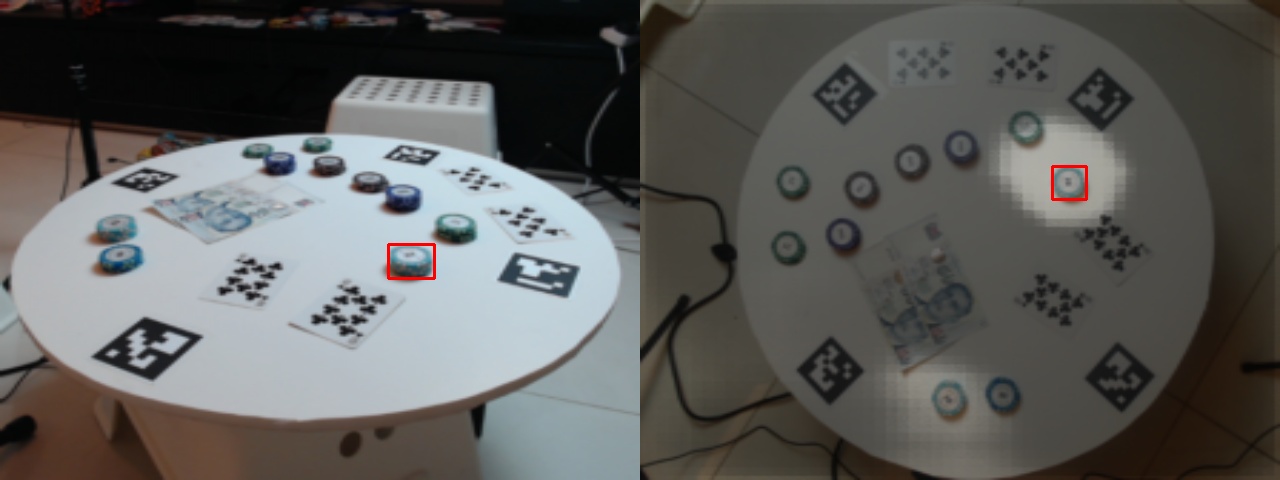}
    \subcaption{Attention mask predicted for a chip stack. Interestingly, two areas in the image are highlighted because there are two identical stacks of chips when viewed from the top. However, the correct chip stack is predicted to have a higher likelihood as it is visualized to be brighter. }
  \end{subfigure}
\caption{The visualization of prediction of attention mask in the right image based on the item in the left image. The red boxes are the ground truth bounding boxes.}
\label{fig:visulization}
\vspace{-5mm}
\end{figure}

To better demonstrate the performance, the predicted mask is superimposed on the full image in input 2:

$$ p_{i,j,c} = p_{i,j,c} \times (0.5 + 0.5 \times \hat{m}_{i,j}) $$
where $c \in \{r, g, b\}$, $i$, $j$ are the row and column indices, $\hat{m}_{i,j} \in (0.0, 1.0)$ is the pixel value in the mask prediction. This computation essentially maps pixel values to 0.5 to 1.0 of its original values based on the mask prediction value that is between 0.0 and 1.0. Visualization result can be seen in Figure \ref{fig:visulization}

In conclusion, the mask module can accurately locate the object of interest in the full image in input 2 by predicting a highly overlapped attention mask. 

\section{Experiments on Association}
We conduct experiments on perfect data without noise except for Section \ref{sec:vs_classical} where noises are added in.

\subsection{Training}

To train the McAssoc and feature-only models, 40 of the 50 scenes from the hard dataset are used in the training set. For each synchronized image pair (top-left or top-right), bounding boxes in the main image can be paired with the bounding box of the same class (e.g. casino chips) in the side image to form the examples. Note bounding boxes of the different class do not participate in the matching as cross-class matching results in too many negative examples and it is unnecessary in real-life application. All training examples are mixed and shuffled to form batches of size 96. 

\subsection{Metrics}

Considering the real-life application of a multi-camera system, we introduce a very strict metric, Image-Pair Association Accuracy (IPAA), to evaluate multi-camera association results.

For each image-pair, only if \textit{all} bounding boxes are associated correctly, the image-pair is counted as correct. This requires an object in one image to be only matched to maximum one object in another, and no ground truth matchings are omitted.

We also report the regular Average Precision (AP) metric as for each bounding box pair, the output association score has a range of 0 to 1 and the label is binary (0 or 1). However, we show in the experiments that IPAA is a more indicative metric: it can be used to differentiate methods that have similar AP.

\subsection{Experiment Results}
\label{sec:er}
We have implemented variations of the 3-branch architecture searching for the optimal network design.In this paper, we adopted three popular neural network architecture (ResNet\cite{ref:resnet}, VGG\cite{ref:vgg}, EfficientNet\cite{ref:en}) as our backbone serves as feature extractor. Experiment results can be found in \ref{table:bkb}

\begin{table}[h]
\begin{center}
\begin{tabular}{|c|c|c|c|}
\hline
Backbone & Params & AP & IPAA \\
\hline\hline
ResNet18 & 11M & 0.996 & 0.9578  \\
VGG16 & 138.36M & 0.996 & 0.9544\\
EfficientNet0 & 5.3M & 0.999 & 0.9756\\
\hline
\end{tabular}
\end{center}
\vspace{-4mm}
\caption{Image-Pair Association Accuracy Results of 3-branch architecture}
\label{table:bkb}
\vspace{-3mm}
\end{table}

Although the feature extraction backbones in each McAssoc module share weights, there is no weight sharing between separate McAssoc modules. In this section, however, we attempt to allow sharing of the weights amongst McAssoc modules to decrease the number of parameters.

\begin{table}[h]
\begin{center}
\begin{tabular}{|c|c|c|c|}
\hline
Backbone & Params & AP & IPAA \\
\hline\hline
ResNet18 & 11M & 0.996 & 0.9511  \\
VGG16 & 138.36M & 0.995 & 0.9322 \\
EfficientNet0  & 5.3M  & 0.999  &0.9771\\
\hline
\end{tabular}
\end{center}
\vspace{-4mm}
\caption{Image-Pair Association Accuracy results of 3-branch architecture with sharing weights}
\vspace{-3mm}
\end{table}

Sharing weights significantly reduce the number of parameters without significantly compromising on the performance. It is viable for systems that require light-weight models.

\subsection{Ablation Studies}
\subsubsection{Branches}
To prove the mask modules help the final prediction of the association score, we compare the 3-branch architecture with a single branch counterpart, that is essentially a single McAssoc module with a head for association score prediction.

\begin{table}[h]
\begin{center}
\begin{tabular}{|c|c|c|c|}
\hline
Architecture & Backbone & AP & IPAA \\
\hline\hline
Single Branch & Res18 & 0.993 & 0.9361  \\
Single Branch & VGG16 & 0.991 & 0.9223 \\
Single Branch & EN0  & 0.995  &0.9542\\
3-Branch & Res18 & 0.996 & 0.9578  \\
3-Branch & VGG16 & 0.995 & 0.9544 \\
3-Branch & EN0  & 0.999  &0.9756\\
\hline
\end{tabular}
\end{center}
\vspace{-4mm}
\caption{Image-Pair Accuracy results of 3-branch architecture with sharing weights. The testing data is rather clean and simple. Refer to Table \ref{tb:noise} for results with noise which is similar to real scenarios.}
\label{tb:noise}
\vspace{-3mm}
\end{table}

It is found out that having the additional information provided by the mask modules (McAssoc modules with mask head), the performance is further boosted.

\subsubsection{Scene understanding and location information}
We carefully design the input to include the full images to allow scene understanding. The attention mask also provides the location information of the object in the entire image. To evaluate this input setup is superior to the intuitive approach where the object is cropped out, we also implement the feature-only approach which simply removes the attention mask. To let the model know which object to match, the object is cropped out and resized to the original image size. 

\begin{table}[h]
\begin{center}
\begin{tabular}{|c|c|c|c|}
\hline
Method & Backbone & IPAA \\
\hline\hline
Feature-only & Res18 & 0.0940  \\
\textbf{McAssoc} & \textbf{Res18} & \textbf{0.9578} \\
\hline
\end{tabular}
\end{center}
\vspace{-3mm}
\caption{Comparison of Image-Pair Association Accuracy between Feature-only and McAssoc}
\vspace{-3mm}
\end{table}

It is worth noting that feature-only approach performs horribly on the strict IPAA metric because the dataset includes item instances that have very similar if not identical appearances.

The obvious limitation of a feature-only approach is that, if there exist many objects with very similar or even the same features (in at least one of the views), they cannot be distinguished if the only feature is used and the matching can fail.

\subsection{Comparison against classical approach}
\label{sec:vs_classical}
In this section, we compare the McAssoc model with the classical approach to different noise types. Refer to Table \ref{noise}, McAssoc outperforms classical approach by a clear margin in terms of error handling.

\begin{table}[h]
\begin{center}
\begin{tabular}{|c|c|c|c|}
\hline
Noise Type & Data Amount & Model/ Method & IPAA \\
\hline\hline
NN & / & Classical & 0.8767  \\
\textbf{NN} & \textbf{8x}  & \textbf{3-Branch} & \textbf{0.9578} \\
FD & / & Classical & 0.3958  \\
\textbf{FD} & \textbf{8x} & \textbf{3-Branch} & \textbf{0.4822} \\
IBB & / & Classical & 0.7967 \\
\textbf{IBB} & \textbf{8x} & \textbf{3-Branch} & \textbf{0.8044} \\
MD  & / & Classical & 0.5067\\
\textbf{MD}  & \textbf{8x} & \textbf{3-Branch} & \textbf{0.9500} \\
All  & /  & Classical & 0.3533\\
All  & 1x  & 3-Branch & 0.3989 \\
All  & 2x  & 3-Branch & 0.4367 \\
All  & 4x  & 3-Branch & 0.5233 \\\
\textbf{All}  & \textbf{8x}  & \textbf{3-Branch} & 0.5356 \\
\hline
\end{tabular}
\end{center}
\vspace{-3mm}
\caption{Comparison of Image-Pair Association Accuracy Results of 3-branch architecture and Classical Approach;\textbf{NN}: No Noise; \textbf{FD}: False Detection; \textbf{IBB}: Imperfect Bounding Box; \textbf{MD}: Missed Detection; \textbf{All}: False Detection + Imperfect Bounding Box + Missed Detection; The definition of \textbf{Data Amount} can be found in Section \ref{sec:data_amount}}
\label{noise}
\vspace{-3mm}
\end{table}

It is worth noting that false detection is the major cause of the drop in IPAA for both classical approach and 3-Branch deep neural network. False detection adds in noisy points that can disturb matching severely. For the deep learning approach, false detection generated might be close to ground truth items that cause confusions.  

Missed detection does not affect the accuracy of McAssoc since it can be interpreted as removing some positive examples only. However, for the classical approach, the missing points can lead to less optimal matching, hurting accuracy. Also, noises can have coupling effects which might accidentally cancel each other.

For the case when all noises are applied (which is the case in real life), it is observed that the deep learning model is performing better than the classical approach even trained with minimum data. As the training set size increases, our solution becomes more advantageous.

\section{Conclusion}
The McAssoc modules and the 3-branch architecture built upon it have been proven to be effective in tackling the multi-camera detection association problem. The proposed deep learning approach does not only perform better than the classical approach but is also more robust when subjected to various noises that are commonly found in real applications. We show in the experiments that merging appearance features and localization information is important for multi-camera association. 

This paper aims to raise research interest in the multi-camera system and McAssoc modules as well as serves as the benchmark for deep learning approach towards cross-camera detection bounding box association.

\section{Appendix}
\subsection{Additional Experiments}
\subsubsection{Realistic (Noisy) Data}
In Section 7 of the paper, all models, except those in Section 7.5, are tested on perfect, noise-free data. Although these comparisons are meant to evaluate the upper limit of the performances these models could achieve, we believe showing their performances on more realistic data (where noises are added) can provide readers with a better view of the strengths of deep learning models over the classical approach.

\begin{table}[h]
\begin{center}
\begin{tabular}{|c|c|c|c|}
\hline
NT & DA & Model/ Method & IPAA \\
\hline\hline
All  & /  & Classical & 0.3533\\
All  & 1x  & 3-Branch(Res18) & 0.3989 \\
All  & 1x  & 3-Branch(EN0) & 0.3978 \\
All  & 1x  & 3-Branch(VGG) & 0.3444 \\
All  & 2x  & 3-Branch(Res18) & 0.4367 \\
All  & 2x  & 3-Branch(EN0) & 0.4356 \\
All  & 2x  & 3-Branch(VGG) & 0.3998 \\
All  & 4x  & 3-Branch(Res18) & 0.5233 \\
All  & 4x  & 3-Branch(EN0) & 0.5244 \\
All  & 4x  & 3-Branch(VGG) & 0.4913 \\
\textbf{All}  & \textbf{8x}  & \textbf{3-Branch(Res18)} & 0.5356 \\
\hline
\end{tabular}
\end{center}
\caption{Comparison of Image-Pair Association Accuracy (IPAA) results of 3-branch architecture (with various backbones and data amount) and classical approach; \textbf{NT}: Noise Type;  \textbf{DA}: Data Amount}
\label{tab:noise}
\end{table}

It is observed in Table \ref{tab:noise} of the Supplementary Material, for various backbones we have tested, even with minimal training data, are generally performing better than the classical approach in terms of the strict metric IPAA. The performances of the models improve substantially when training data increases. Amongst three backbones, VGG-16 is the bulkiest, yet not as good as the other two. This can be attributed to the fact that VGG-16 is a relatively old design. In contrast, the newer ResNet-18 and the recently released EfficientNet performs even better with only a fraction of the number of parameters.

\subsubsection{Realistic (Noisy) Data with Post-Processing}

The loss used in the classical approach is pixel distance between two points, for which finding a suitable threshold value to separate a match and a non-match is difficult. Hence, the classical approach has to include a post-processing stage to assign matches by optimizing globally.

Our deep learning models, however, output association score that falls within a range between 0 and 1. Hence, it is intuitive to set a threshold of 0.5 to determine if two points are matched. However, having all the matches optimized globally will surely increase the chance of getting all the matches right: for an item in one view, the network might give high association scores to multiple items in the other view that all look similar but there should be at most only one match; optimizing globally can reject those with relatively low scores.   

In Table 3 of the Supplementary Material, it is observed that even without post-processing, the deep learning models are significantly better than the classical approach. When added post-processing in the evaluation stage, notable improvements (as large as 0.2 or 20\%) are observed.

Note that post-processing is added only in evaluation; the training is end-to-end;

\begin{table}[h]
\begin{center}
\begin{tabular}{|c|c|c|c|}
\hline
NT & DA & Model/ Method & IPAA \\
\hline\hline
All  & /  & Classical & 0.3533\\
All  & 4x  & 3-Branch(Res18) & 0.5233 \\
All  & 4x  & 3-Branch(Res18) + KM & 0.6556 \\
All  & 4x  & 3-Branch(EN0) & 0.5244 \\
All  & 4x  & 3-Branch(EN0) + KM & 0.7511 \\
All  & 4x  & 3-Branch(VGG) & 0.4913 \\
All  & 4x  & 3-Branch(VGG) + KM & 0.6021 \\
All  & 8x  & 3-Branch(Res18) & 0.5356 \\
\textbf{All} & \textbf{8x}  & \textbf{3-Branch(Res18) + KM} & \textbf{0.7556} \\
\hline
\end{tabular}
\end{center}
\caption{Comparison of Image-Pair Association Accuracy (IPAA) results of 3-branch architecture (with post-processing adopting KM Algorithm) and classical approach}
\label{tab:post-proc}
\end{table}

\subsection{Additional Explanations}
We provide more detailed explanations on the use of Kuhn-Munkres algorithm and the effects of imperfect bounding boxes on the results of the classical approach.

\subsubsection{Kuhn-Munkres algorithm}
Kuhn-Munkres algorithm is used for maximum bipartite matching with minimum loss. That is, give two sets of vertices, we hope to draw a maximum number of edges, while the sum of losses associated with each edge is minimized. In our case, the loss is the pixel distance between vertices: the further away the vertices are from each other, the larger the loss. 

The output of the Kuhn-Munkres algorithm gives an assignment matrix that has the same shape as the cost matrix. In the assignment matrix, values at row i and column j can only be 1 or 0, indicating a match or a non-match between bounding box i in image 1 and bounding box j in image 2. The assignment satisfies that one bounding box from one image can only be matched to maximum one bounding box in the other image (there are maximum one "1" in each row and each column), also the sum of all losses of assignments are minimal.

\subsubsection{Impacts of False or Missed Detection on Classical Approach}
\begin{figure}[h]
  \begin{subfigure}[b]{\columnwidth}
    \centering\includegraphics[width=0.75\columnwidth]{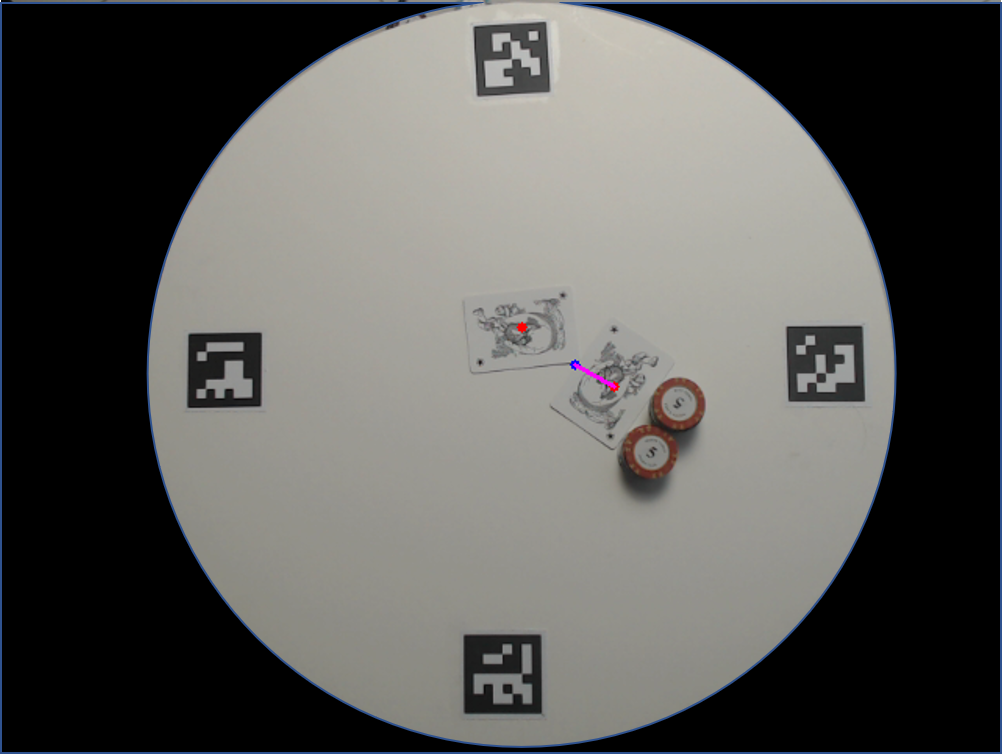}
    \caption{A simple case where the playing card in the right cannot be seen by the side-view camera (in the bottom right corner) because of occlusion (two stacks of chips). The correct match should be between the blue point and the red point on the left. The red point on the right should not have any match as its corresponding blue point does not exist due to missed detection in the side view.}
    \label{fig:md_a}
  \end{subfigure}
  \begin{subfigure}[b]{\columnwidth}
    \centering\includegraphics[width=0.75\columnwidth]{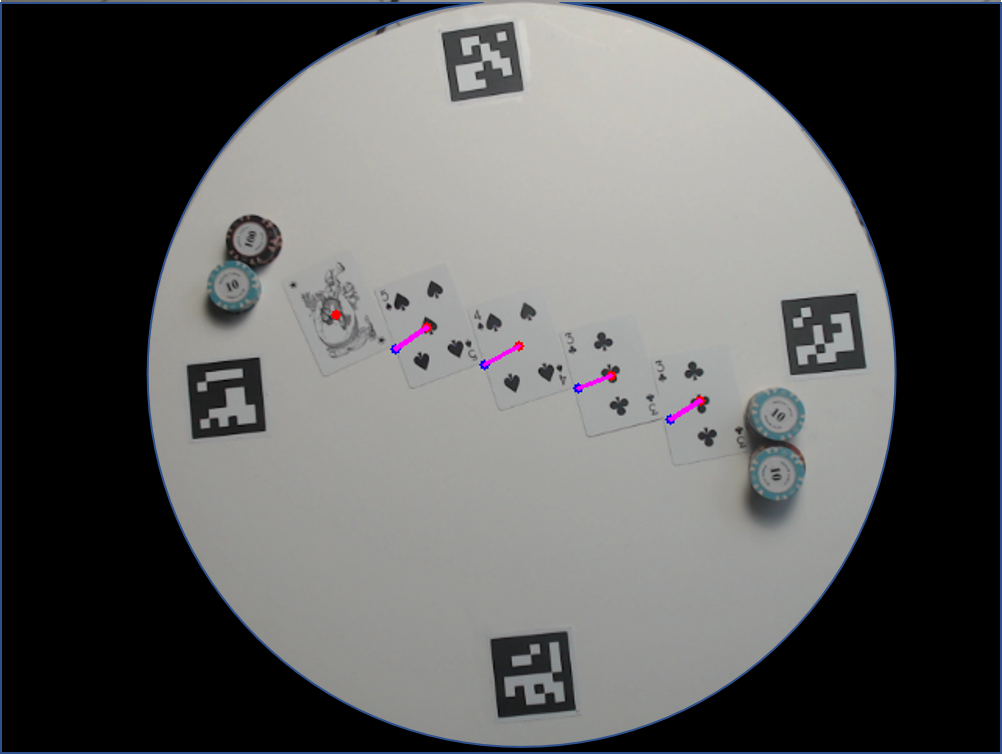}
    \subcaption{The same case in Figure \ref{fig:md_a} can be extended in a more extreme case where the missed detection can result in a sequence of wrong matches.}
  \end{subfigure}
  \begin{subfigure}[b]{\columnwidth}
    \centering\includegraphics[width=0.75\columnwidth]{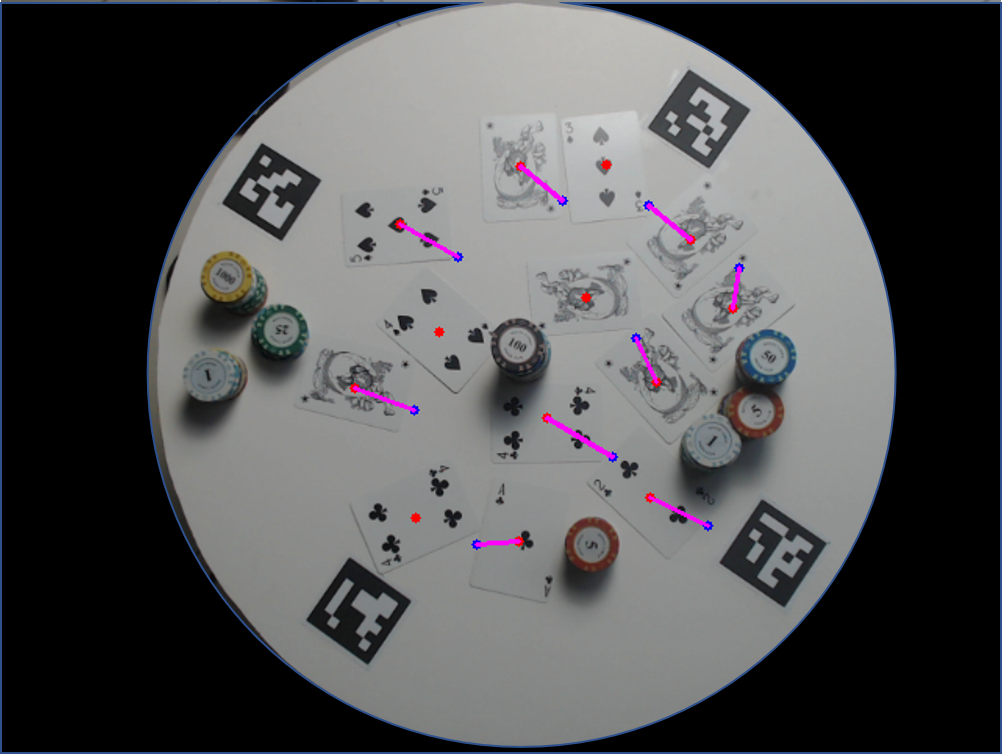}
    \subcaption{A more realistic case showing missed detection can result in many wrong matches. Correct match should have the match lines approximately pointing at the bottom right corner.}
  \end{subfigure}
\caption{Visualization of missed detection's impact on classical approach. }
\label{fig:md}
\end{figure}

In Section 7.5 of the paper, we evaluate the performance drops due to noises in detection. In this section (Figure \ref{fig:fd} and Figure \ref{fig:md}, we explain the classical approach' sensitivity to false and missed detection with visualization.

\begin{figure}[h]
  \begin{subfigure}[b]{\columnwidth}
    \centering\includegraphics[width=0.75\columnwidth]{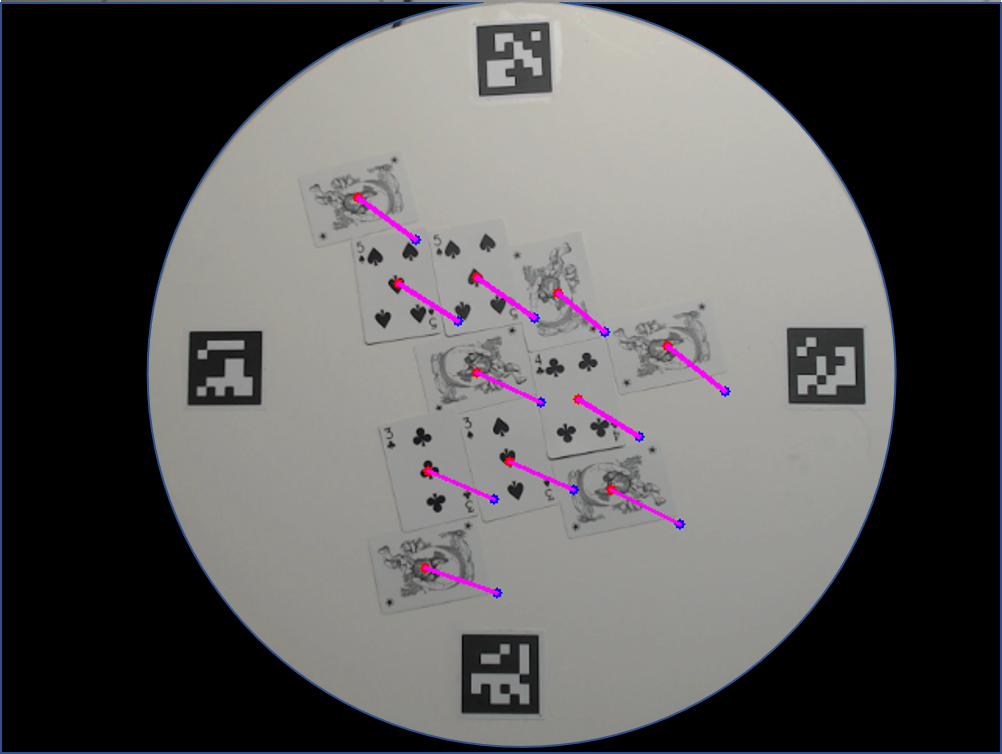}
    \caption{Correct match when no false detection is added.}
  \end{subfigure}
  \begin{subfigure}[b]{\columnwidth}
    \centering\includegraphics[width=0.75\columnwidth]{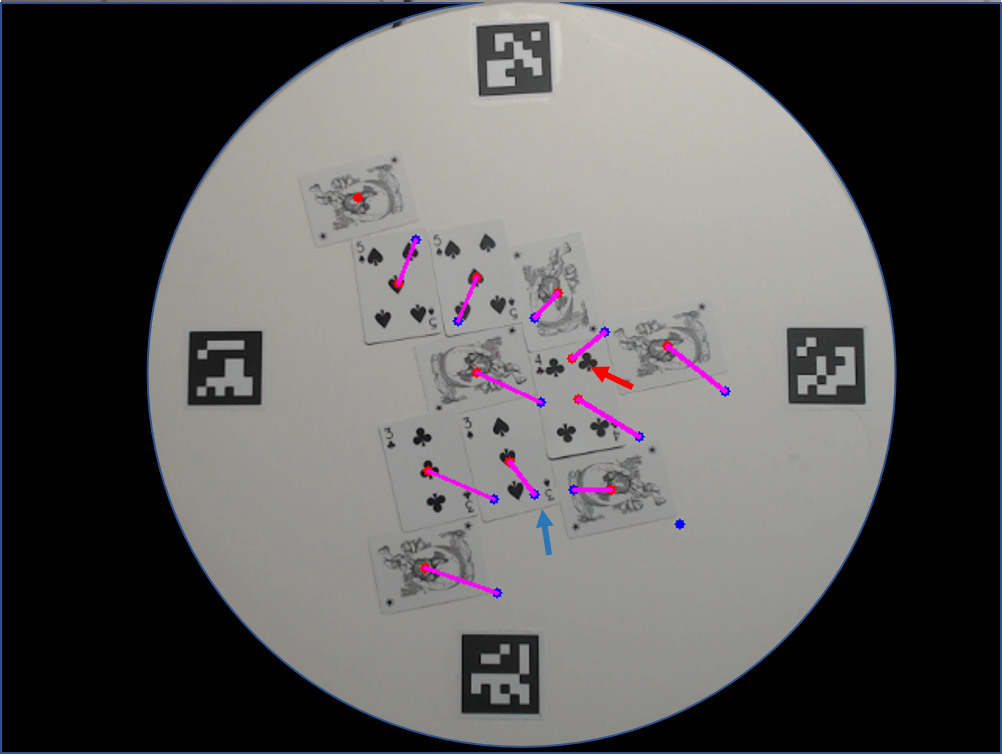}
    \subcaption{Incorrect match with two random false detection bounding boxes, one in the bird view (indicated by the red arrow) and one in the side view, projected to the bird view (indicated by the blue arrow). Merely one false detection in each camera can result in serious errors.}
  \end{subfigure}
\caption{Visualization of false detection's impact on the classical approach. Red points are the center point of the bounding boxes in the bird view, blue points are the mid-points of the bottom edges of the bounding boxes in the side view camera, projected to the bird view. Magenta lines indicate matches between two points. The side camera is installed at the bottom right corner when viewed in the bird view, hence, a correct match line should point at the bottom right corner of the image.}
\label{fig:fd}
\end{figure}

\subsection{Dataset}
We visualize some synchronized images with all annotated bounding boxes from the dataset. The index of the bounding boxes are indicated as numbers in the boxes. Bounding boxes with the same index indicate association.

\begin{figure}[h]
  \begin{subfigure}[b]{\columnwidth}
    \centering\includegraphics[width=\columnwidth]{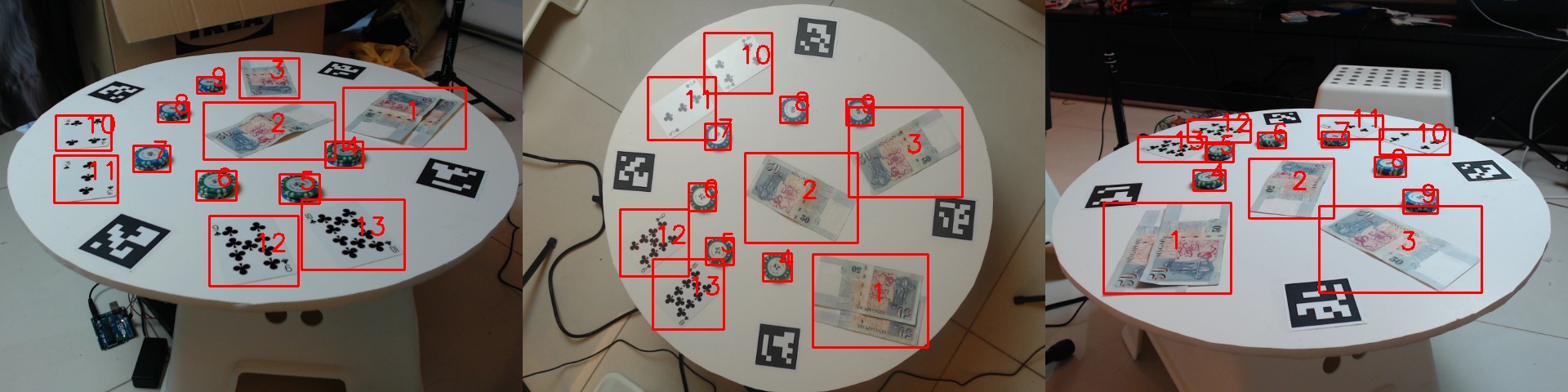}
    \caption{A mixture of cash, chip and playing cards.}
  \end{subfigure}
  \begin{subfigure}[b]{\columnwidth}
    \centering\includegraphics[width=\columnwidth]{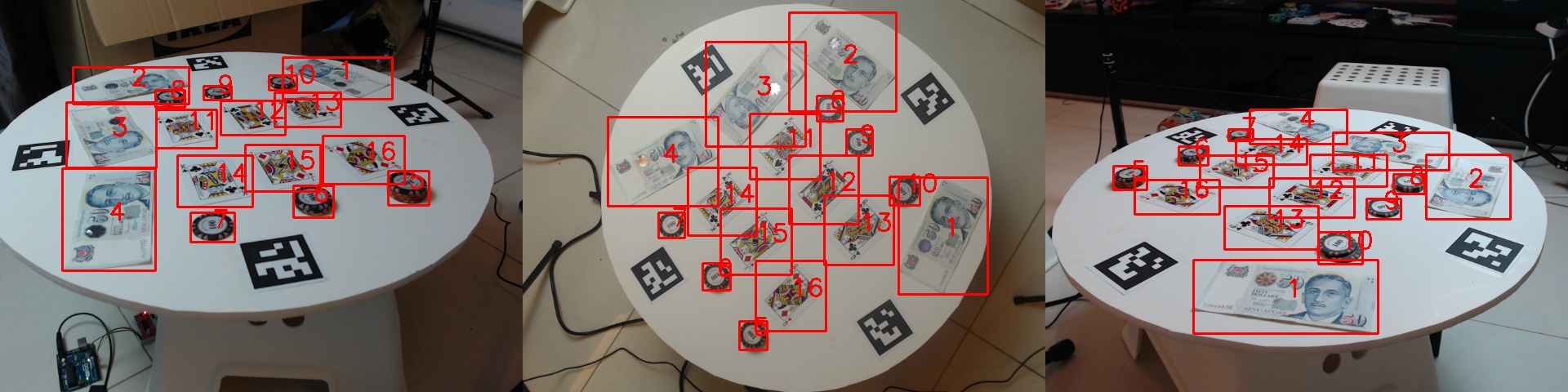}
    \subcaption{A denser mixture of cash, chip and playing cards.}
  \end{subfigure}
  \begin{subfigure}[b]{\columnwidth}
    \centering\includegraphics[width=\columnwidth]{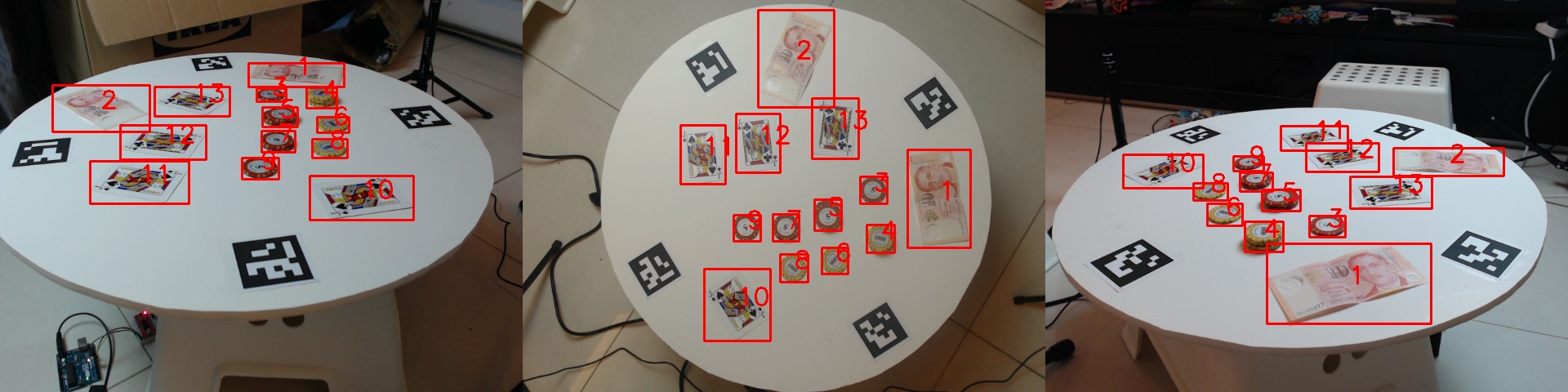}
    \subcaption{Contains arrays of similar chip stacks.}
  \end{subfigure}
    \begin{subfigure}[b]{\columnwidth}
    \centering\includegraphics[width=\columnwidth]{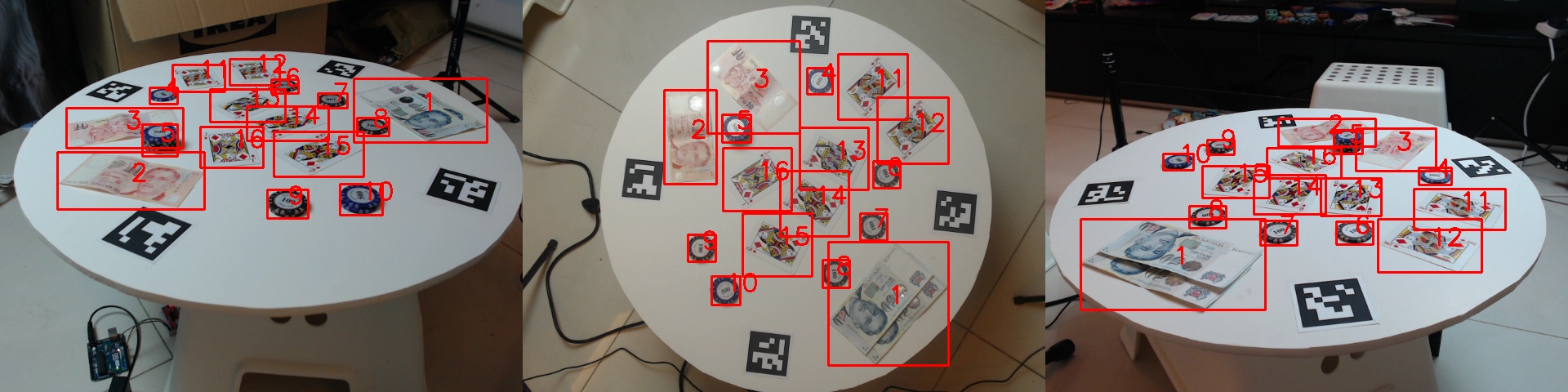}
    \subcaption{Contains similar playing cards.}
  \end{subfigure}
    \begin{subfigure}[b]{\columnwidth}
    \centering\includegraphics[width=\columnwidth]{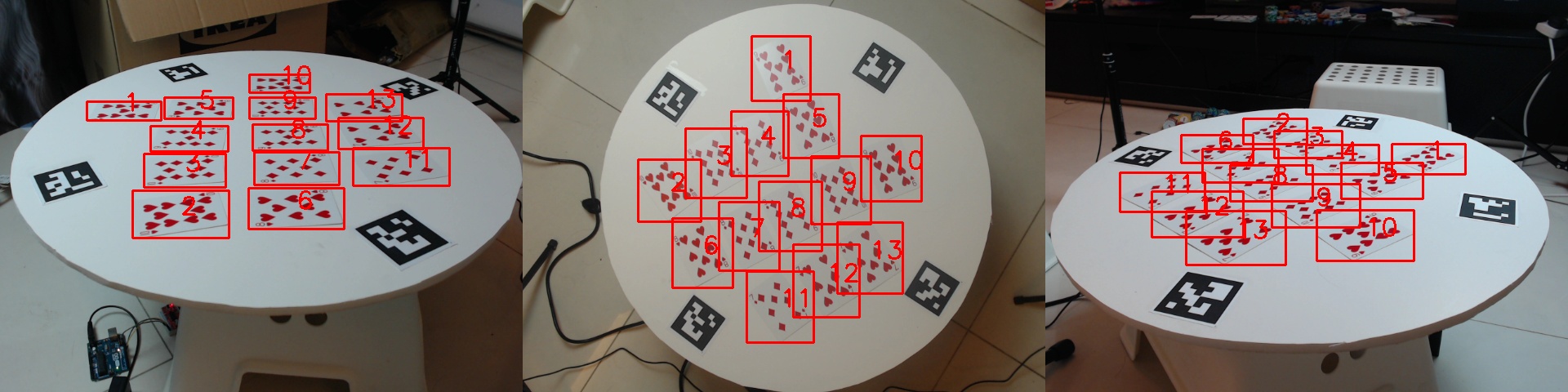}
    \subcaption{Only similar playing cards.}
  \end{subfigure}
    \begin{subfigure}[b]{\columnwidth}
    \centering\includegraphics[width=\columnwidth]{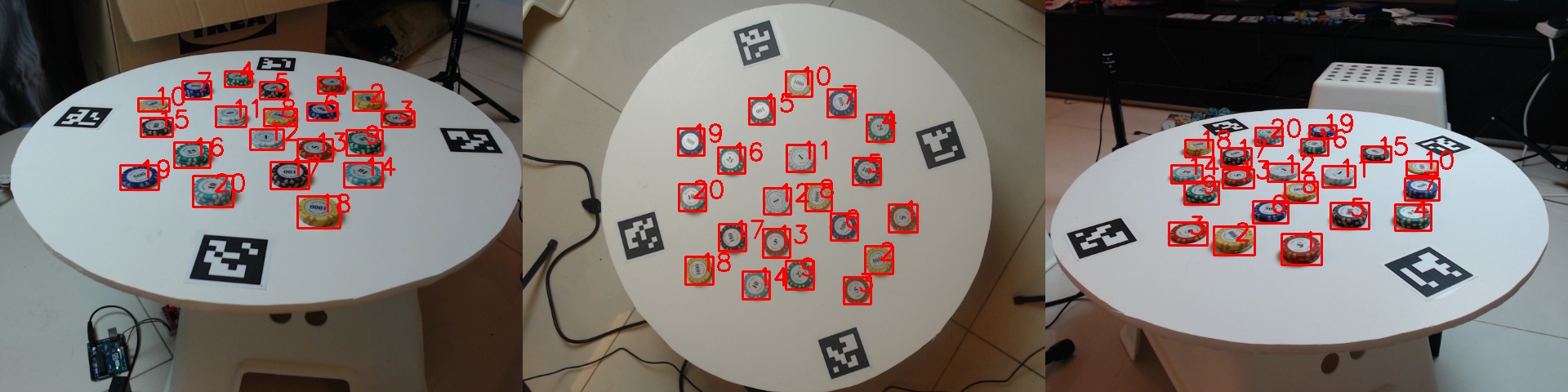}
    \subcaption{Only similar chip stacks.}
  \end{subfigure}
\caption{Visualization of some synchronized image sets in the dataset. Each set contains a left view, a bird view and a right view. We purposely added in many similar-looking instances to confuse the neural networks. Also, items are placed in dense cluster to make it more challenging.}
\label{fig:fd}
\end{figure}

{\small
\bibliographystyle{ieee}
\bibliography{reference}
}

\end{document}